\pdfoutput=1

\documentclass[11pt]{article}

\usepackage[]{acl}

\usepackage{times}
\usepackage{latexsym}
\usepackage{amsfonts}
\usepackage{amsmath}
\usepackage[ruled,linesnumbered]{algorithm2e}
\usepackage{algpseudocode}
\usepackage{makecell}
\usepackage{booktabs}
\usepackage{multirow}
\usepackage{graphicx} 
\usepackage{float} 
\usepackage{subfigure} 
\newcommand\ourmodel{FPT\xspace}
\usepackage{bm}
\usepackage{color}
\usepackage{diagbox}
\usepackage[T1]{fontenc}

\usepackage[utf8]{inputenc}

\usepackage{microtype}

\usepackage{cleveref}
\crefname{section}{§}{§§}
\Crefname{section}{§}{§§}

\title{FPT: Improving Prompt Tuning Efficiency via Progressive Training}

\author{
 Yufei~Huang$^{1,2,3}\thanks{\ \ Indicates equal contribution.}$\hspace{0.5em}, Yujia~Qin$^{1,2,3*}$, Huadong~Wang$^{1,2,3}$, Yichun~Yin$^{4}$, \\ \textbf{Maosong~Sun$^{1,2,3,5}\thanks{\ \  Corresponding author.}$\hspace{0.5em}, Zhiyuan~Liu$^{1,2,3\dag}$, Qun~Liu$^4$} \\
 $^1$DCST, Tsinghua University, Beijing $^2$BNRIST, Tsinghua University, Beijing \\
 $^3$IAI, Tsinghua University, Beijing $^4$Huawei Noah’s Ark Lab \\
 $^5$Jiangsu Collaborative Innovation Center for Language Ability, Xuzhou, China \\
\texttt{\{huang-yf20, qyj20\}@mails.tsinghua.edu.cn}\\
}

\begin{document}
\maketitle

\begin{abstract}
Recently, prompt tuning (PT) has gained increasing attention as a parameter-efficient way of tuning pre-trained language models (PLMs). Despite extensively reducing the number of tunable parameters and achieving satisfying performance, PT is training-inefficient due to its slow convergence. To improve PT's training efficiency, we first make some novel observations about the prompt transferability of ``partial PLMs'', which are defined by compressing a PLM in depth or width. We observe that the soft prompts learned by different partial PLMs of various sizes are similar in the parameter space, implying that these soft prompts could potentially be transferred among partial PLMs. Inspired by these observations, we propose Fast Prompt Tuning (FPT), which starts by conducting PT using a small-scale partial PLM, and then progressively expands its depth and width until the full-model size. After each expansion, we recycle the previously learned soft prompts as initialization for the enlarged partial PLM and then proceed PT. We demonstrate the feasibility of \ourmodel on $5$ tasks and show that \ourmodel could save over $30$\% training computations while achieving comparable performance. The codes are publicly available at \url{https://github.com/thunlp/FastPromptTuning}.
\end{abstract}
\section{Introduction}
The emergence of pre-trained language models (PLMs) has broken the glass ceiling for various NLP tasks~\citep{han2021pre}. Versatile semantic and syntactic knowledge acquired during pre-training could be leveraged when PLMs are adapted towards a specific downstream task to boost performance. The de facto strategy for such an adaptation is full-parameter fine-tuning, which is computationally expensive and profligate since it requires tuning and storing all the parameters in the PLM for each downstream task.
\begin{figure}[!t]
    \centering
    \includegraphics[width=0.48\textwidth]{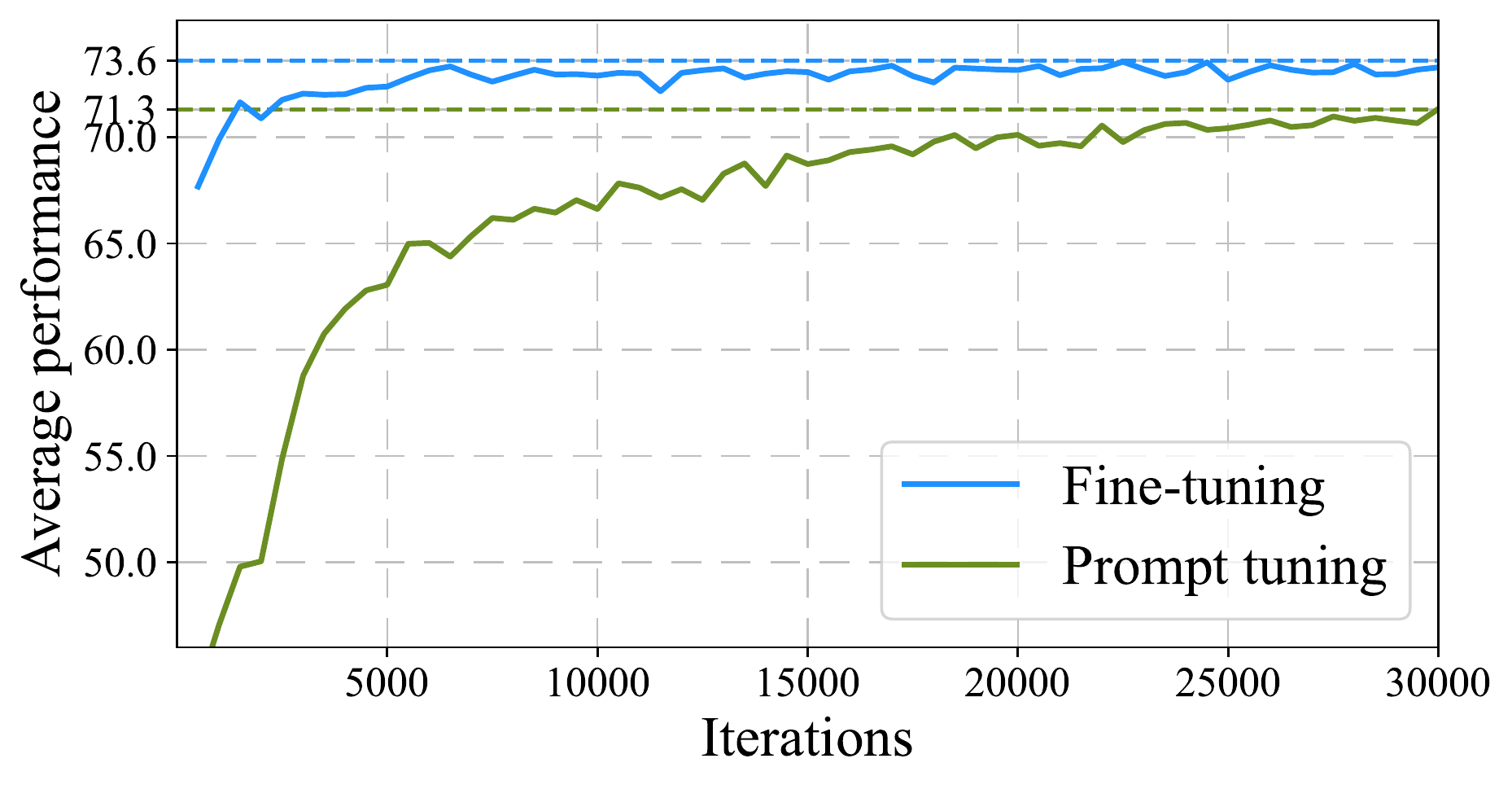} 
    \caption{Average performance growth of $\text{T5}_{\texttt{LARGE}}$ on $5$ investigated tasks in this paper comparing fine-tuning and PT. The convergence speed of PT is much slower than fine-tuning in terms of training steps.}
    \label{fig:finetuning_prompttuning}
\end{figure}
To remedy this, several delta tuning~\citep{ding2022delta} (also known as parameter-efficient tuning) algorithms are proposed in place of the vanilla fine-tuning~\citep{houlsby2019parameter,li-liang-2021-prefix,hu2021lora,zaken2021bitfit}, among which prompt tuning (PT)~\citep{lester-etal-2021-power} has gained increasing attention recently. PT prepends a few \textit{virtual tokens} to the input text, these tokens are tuned during training while all the other PLM parameters remain frozen. Despite its simple form, PT has been demonstrated to achieve remarkable performance in various NLP tasks. Especially when the scale of the PLM becomes extremely huge, PT could achieve comparable performance to fine-tuning~\citep{lester-etal-2021-power}. Despite extensively reducing the number of tunable parameters and achieving satisfying performance, PT is criticized to be training-inefficient due to the slow convergence \citep{su2021transferability} as illustrated in Figure~\ref{fig:finetuning_prompttuning}, and such incompetence would limit the practical application of PT. Hence in this paper, we explore how to improve PT's training efficiency.

\begin{figure}[!t]
    \centering
        \includegraphics[width=0.48\textwidth]{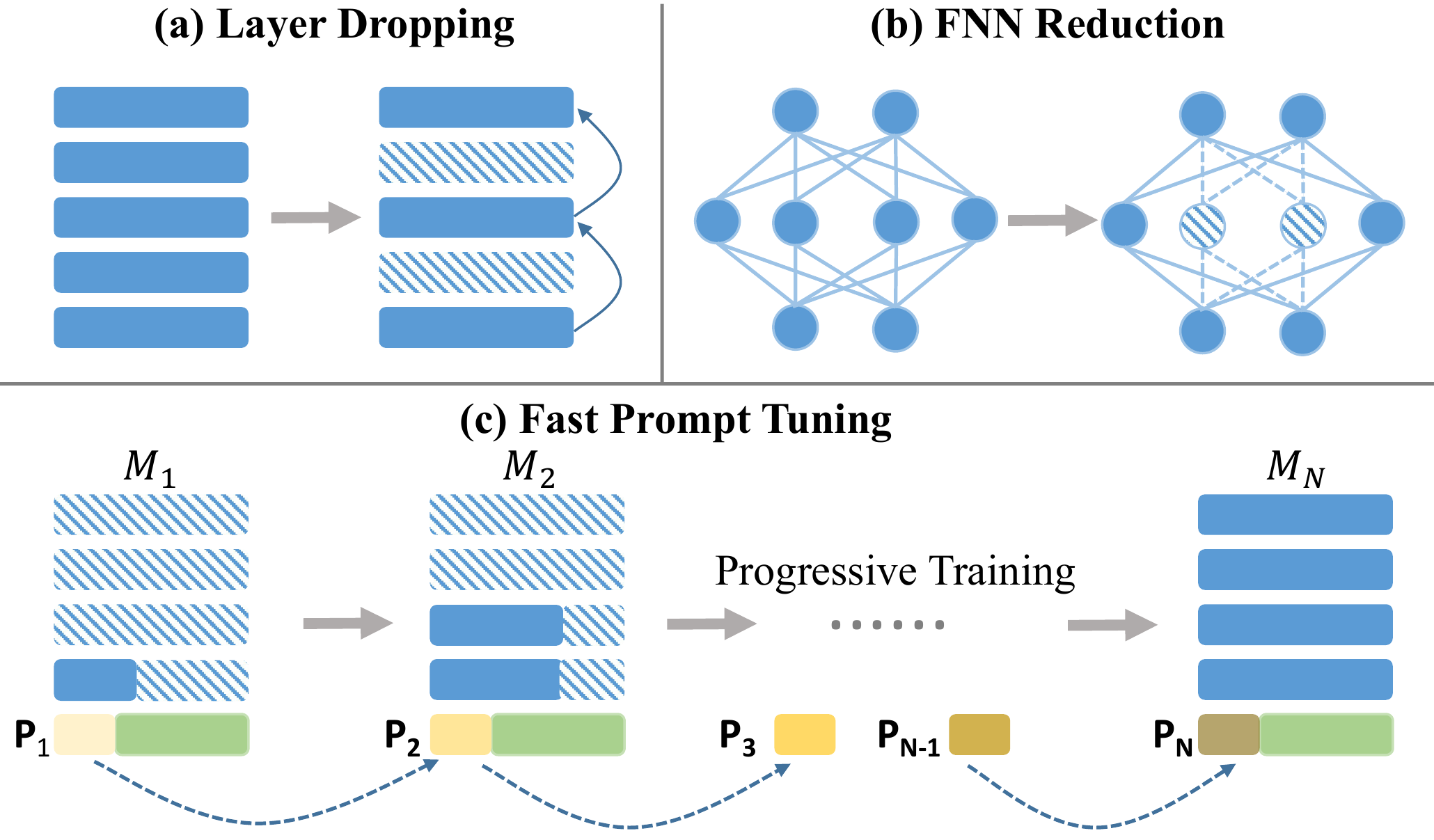}
    \caption{The framework of Fast Prompt Tuning (FPT). The top part (a,b) shows two methods to construct a partial PLM. The bottom part (c) shows FPT's training process, we conduct PT on a partial PLM, progressively expand its size and transfer the trained prompts.} 
    \label{fig:model} 
\end{figure}

Our motivation is based on novel observations about the prompt transferability among ``partial PLMs''. Here a partial PLM is defined by compressing a PLM in depth or width, which is implemented by dropping several layers or masking part of the connections in the feed-forward network (FFN) in each Transformer~\citep{vaswani2017attention} layer. We observe that the soft prompts of the same task learned by different partial PLMs of various sizes tend to be close in the parameter space, implying that these soft prompts could potentially be transferred among different partial PLMs.

Inspired by the above observations, we propose Fast Prompt Tuning (FPT), which starts by conducting PT using a small-scale partial PLM to obtain the corresponding soft prompts. After that, we progressively expand the partial PLM's depth and width until the full-model size by rehabilitating the dropped layers and masked neurons. After each expansion, we recycle the previously learned soft prompts as initialization for the enlarged PLM and then proceed PT. Since the partial PLM requires fewer computations for each step, keeping the total training steps unchanged, we could reduce the overall computations consumed, and in the meantime, achieve comparable PT performance. In experiments, we demonstrate the feasibility of \ourmodel on $5$ NLP tasks. The experimental results show that \ourmodel could save around $30$\% training computations and achieve satisfying downstream performance.

\section{Prompt Tuning on a Partial PLM}
\label{sec:preliminary}

\subsection{Prompt Tuning}
For a given input sequence $\mathcal{X} = \{x_1, x_2, ... ,x_{n}\}$ and its target label $\mathcal{Y}$, PT first converts $\mathcal{X}$ into a matrix $\mathbf{X} \in \mathbb{R} ^ {n \times d}$, where $d$ is the hidden size. After that, PT prepends $l$ tunable soft prompt tokens $\mathbf{P} \in \mathbb{R} ^{l \times d}$ before $\mathbf{X}$, creating a new input matrix $[\mathbf{P};\mathbf{X}] \in \mathbb{R} ^{(l + n) \times d}$, which is then processed by the PLM. The training objective is to maximize $\mathcal{P}(\mathcal{Y}|[\mathbf{P};\mathbf{X}])$, where only $\mathbf{P}$ is optimized during training and the parameters of PLM are frozen. 
Although PT is applied to the entire PLM by default, in this section, we investigate how the performance would become if we conduct PT on a partial PLM, i.e., only part of the parameters in the PLM participate in the computation.

\subsection{Partial PLM Construction}
\label{subsec:partial_model_construction}
Using partial parameters in a PLM is typically applied to reduce the \textbf{inference computation} for fine-tuning, such as early exit~\citep{teerapittayanon2016branchynet,xin2020deebert} and model pruning~\citep{chen2020lottery, sun-etal-2020-mobilebert, Fan2020Reducing}, which assume that the features produced by a part of a PLM may already suffice to classify some input examples. In this paper, we investigate its application in reducing the \textbf{training computation} of PT, and propose to construct partial PLMs by shrinking the original PLM in both depth and width, as illustrated in Figure~\ref{fig:model} (a, b). Details are listed in \cref{sec:training_details}.

\begin{table*}[!t]
    \centering
    \small
    \begin{tabular}{l|c@{~~~}c@{~~~}|c@{~~~}c@{~~~}c@{~~~}c@{~~~}c@{~~~}|c@{~~~}c@{~~~}c@{~~~}c}
        \toprule[1pt]
        \ & \textbf{Enc./Dec.} & \textbf{FFN} & $\textsc{MNLI}$ & $\textsc{QQP}$ & $\textsc{SQuAD2.0}$ & $\textsc{ReCoRD}$ & $\textsc{XSum}$ & \multirow{2}*{Avg.} & \multirow{2}*{$\Delta$} & \multirow{2}*{FLOPs} & Wall\\
        & \textbf{Layer} & \textbf{Dimension} & (Acc) & (Acc) & (EM) & (EM) & (ROUGE-L) & & & & Clock \\
        \midrule[1pt]
        \textbf{PT} & 24 / 24 & 2,816 & \textbf{86.07} & \textbf{87.26} & \textbf{76.09} & \textbf{81.46} & \textbf{26.65} & \textbf{71.51} & - & 100\% & 100\% \\
        \hline
        \multirow{3}{*}{\shortstack[l]{\textbf{LD}}}
        & 6 / 6 & 2,816 & 60.34 & 78.29 & 48.14 & 24.75 & 17.40 & 45.78 & -25.73 & 30\% & 35\% \\
        & 12 / 12 & 2,816 & 63.90 & 80.64 & 52.87 & 39.09 & 19.69 & 51.24 & -20.27 & 54\% & 56\% \\
        & 18 / 18 & 2,816 & 81.41 & 86.05 & 63.97 & 59.87 & 22.51 & 62.76 & -8.75 & 77\% & 77\%\\
        \hline
        \multirow{3}{*}{\shortstack[l]{\textbf{FR}}} 
        & 24 / 24 & 704 & 78.18 & 85.19 & 66.68 & 62.90 & 22.46 & 63.08 & -8.43 & 58\% & 72\%\\
        & 24 / 24 & 1,408 & 82.62 & 86.45 & 72.65 & 74.59 & 24.61 & 68.19 & -3.32 & 72\% & 81\% \\
        & 24 / 24 & 2,112 & \underline{84.93} & \underline{86.77} & \underline{74.73} & \underline{79.52} & \underline{26.12} & \underline{70.41} & \underline{-1.10} & 86\% & 91\% \\
        \hline
        \multirow{3}{*}{\shortstack[l]{\textbf{CR}}}
        & 6 / 6 & 704 & 62.53 & 78.38 & 48.62 & 23.99 & 16.49 & 46.00 & -25.51 & 20\% & 28\% \\
        & 12 / 12 & 1,408 & 64.09 & 78.91 & 50.90 & 29.50 & 18.88 & 48.45 & -23.06 & 40\% & 47\% \\
        & 18 / 18 & 2,112 & 80.63 & 86.32 & 63.42 & 58.97 & 22.18 & 62.30 & -9.21 & 66\% & 71\%\\
        \bottomrule[1pt]
    \end{tabular}
    \caption{Average results for partial PLM PT on T5$_{\texttt{LARGE}}$ with layer dropping (\textbf{LD}), FFN Reduction (\textbf{FR}), and compound reduction (\textbf{CR}). $\Delta$ denotes the performance degeneration compared with vanilla \textbf{PT} of each setting. The ``FLOPs'' and ``Wall Clock'' columns are both relative values compared with \textbf{PT} and are averaged over $5$ tasks.}
    \label{tab:preliminary_results}
\end{table*}

\paragraph{Layer Dropping.}
Based on previous findings \citep{clark-etal-2019-bert,jawahar-etal-2019-bert} that adjacent layers in PLMs generally have similar functionalities, we hypothesize that removing part of these layers may not significantly hurt the overall performance, and we propose to drop a PLM's layers uniformly to construct a partial PLM consisting of fewer layers than the original PLM. After that, we directly build connections among the remaining layers keeping the original order, which is found empirically to work well although such connections do not exist during pre-training.


\paragraph{FFN Reduction.}
\citet{jaszczur2021sparse} and \citet{zhang2021moefication} indicate that only part of the neurons in the FFN layers will be activated for a given input. Such a sparse activation phenomenon inspires us to reduce the computation in FFN by shrinking the width of the FFN layer. Specifically, the FFN layer consists of two fully connected networks with a nonlinear activation function $\sigma$, and it processes an input representation $\bm{x}\in \mathbb{R}^{d}$ as: $ \mathrm{FFN}(\bm{x}) = \sigma(\bm{x}\bm{W}_1+\bm{b}_1)\bm{W}_2+\bm{b}_2$, where $\bm{W}_1\in \mathbb{R}^{d\times d'} $ and $\bm{W}_2\in \mathbb{R}^{ d'\times d}$ are the weight matrices, $\bm{b}_1\in \mathbb{R}^{d'} $ and $\bm{b}_2\in \mathbb{R}^{d}$  are the bias terms. We abandon a portion of $\bm{W}_1$ / $\bm{W}_2$'s columns / rows (i.e., reducing $d'$) by masking the neurons that are seldom activated. In practice, before training, we feed a few downstream examples prepended by randomly initialized soft prompts into the full-size PLM and record the neuron activation of each dimension of $d'$.
\paragraph{Compound Reduction.}
Since the above methods are compatible with each other, we try to combine them to form a partial PLM smaller than the original PLM in both depth and width.

\subsection{Observations}
\label{subsec:partial_model_experiment}


To explore PT's performance on a partial PLM, we conduct experiments on $\text{T5}_{\texttt{LARGE}}$ \citep{2020t5}. We choose $5$ representative NLP datasets in English, covering the tasks of natural language inference (\textsc{MNLI}~\citep{williams-etal-2018-broad}), paraphrase (\textsc{QQP}~\href{https://quoradata.quora.com/First-Quora-Dataset-Release-Question-Pairs}{(link)}), reading comprehension (\textsc{SQuAD2.0}~\citep{rajpurkar-etal-2018-know} and \textsc{ReCoRD}~\citep{zhang2018record}), and summarization (\textsc{XSum}~\citep{narayan-etal-2018-dont}). For both layer dropping and FFN reduction, we evaluate the performance when we reduce the number of Transformer layers or FFN intermediate dimension to $\{\frac{1}{4}, \frac{1}{2}, \frac{3}{4}\}$. We train all models using the same steps and the details are described in \cref{sec:training_details}.

\begin{table*}[!t]
    \centering
    \small
    \begin{tabular}{l|l|ccccc|c|c|c|c}
        \toprule[1pt]
        \multicolumn{2}{c|}{\multirow{2}{*}{\textbf{Method}}} & $\textsc{MNLI}$ & $\textsc{QQP}$ & $\textsc{SQuAD2.0}$ & $\textsc{ReCoRD}$ & $\textsc{XSum}$ & \multirow{2}*{Avg.} & \multirow{2}*{FLOPs} & Wall & \multirow{2}*{Improve$\uparrow$} \\
        \multicolumn{2}{l|}{\ } & (Acc) & (Acc) & (EM) & (EM) & (ROUGE-L) & & & Clock & \\
        \midrule[1pt]
        \multirow{4}*{T5$_{\texttt{LARGE}}$}
        & PT & 86.07 & \textbf{87.26} & 76.09 & \textbf{81.46} & \textbf{26.65} & \textbf{71.51} & 100\%  & 100\% & -\\
        & FPT$_{\text{LD}}$ & 85.72 & 86.51 & 75.89 & 80.23 & 26.27 & 70.92 & 72\% & 74\% & 0.11\\
        & FPT$_{\text{FR}}$ & \textbf{86.49} & 87.11 & \textbf{76.26} & 81.07 & 26.55  & 71.50 & 83\% & 89\% & \textbf{0.49} \\
        & FPT$_{\text{CR}}$ & 85.13 & 86.40 & 75.63 & 81.00 & 26.21 & 70.87 & \textbf{65\%} & \textbf{70\%} & 0.38 \\
        \midrule[1pt]
        \multirow{4}*{T5$_{\texttt{XL}}$}
        & PT & 89.00 & 88.20 & 81.08 & \textbf{88.48} & \textbf{30.53} & 75.46 & 100\%  & 100\% & -\\
        & FPT$_{\text{LD}}$ & 88.99 & 88.09 & \textbf{82.18} & 88.06 & 30.52 & \textbf{75.57} & 86\% & 86\% & \textbf{0.78} \\
        & FPT$_{\text{FR}}$ & 88.84 & \textbf{88.21} & 81.74 & 88.46 & 30.52 & 75.55 & 84\% & 87\% & 0.76 \\
        & FPT$_{\text{CR}}$ & \textbf{89.18} & 87.34 & 80.88 & 87.82 & 30.43 & 75.13 & \textbf{74\%} & \textbf{76\%} & 0.48 \\
        \bottomrule[1pt]
    \end{tabular}
    \caption{Performance of the vanilla PT and three variants of our method. FPT$_{\text{LD}}$, FPT$_{\text{FR}}$, and FPT$_{\text{CR}}$ refer to constructing partial PLMs by layer dropping, FFN reduction, and compound reduction. The column ``Improve$\uparrow$'' denotes the performance improvement of each FPT$_{*}$ method over PT when PT uses the same FLOPs as FPT$_{*}$.}
    \label{tab:results}
\end{table*}

\paragraph{Overall Performance.}
The overall results are shown in Table~\ref{tab:preliminary_results}. We observe that for each method, \textbf{despite abandoning a large portion of parameters, a partial PLM reserves most of the PT performance of the full-size PLM}. As expected, the performance becomes better when more parameters are retained. In addition, we find that the performance drop is less sensitive to FFN reduction than layer dropping. Specifically, there is only $1.10$\% performance drop on average when $25$\% neurons are masked. These results indicate that the resulting partial PLM still retains most of the functionalities of the original PLM.

\begin{figure}[t]
    \centering
    \includegraphics[width=0.30\textwidth]{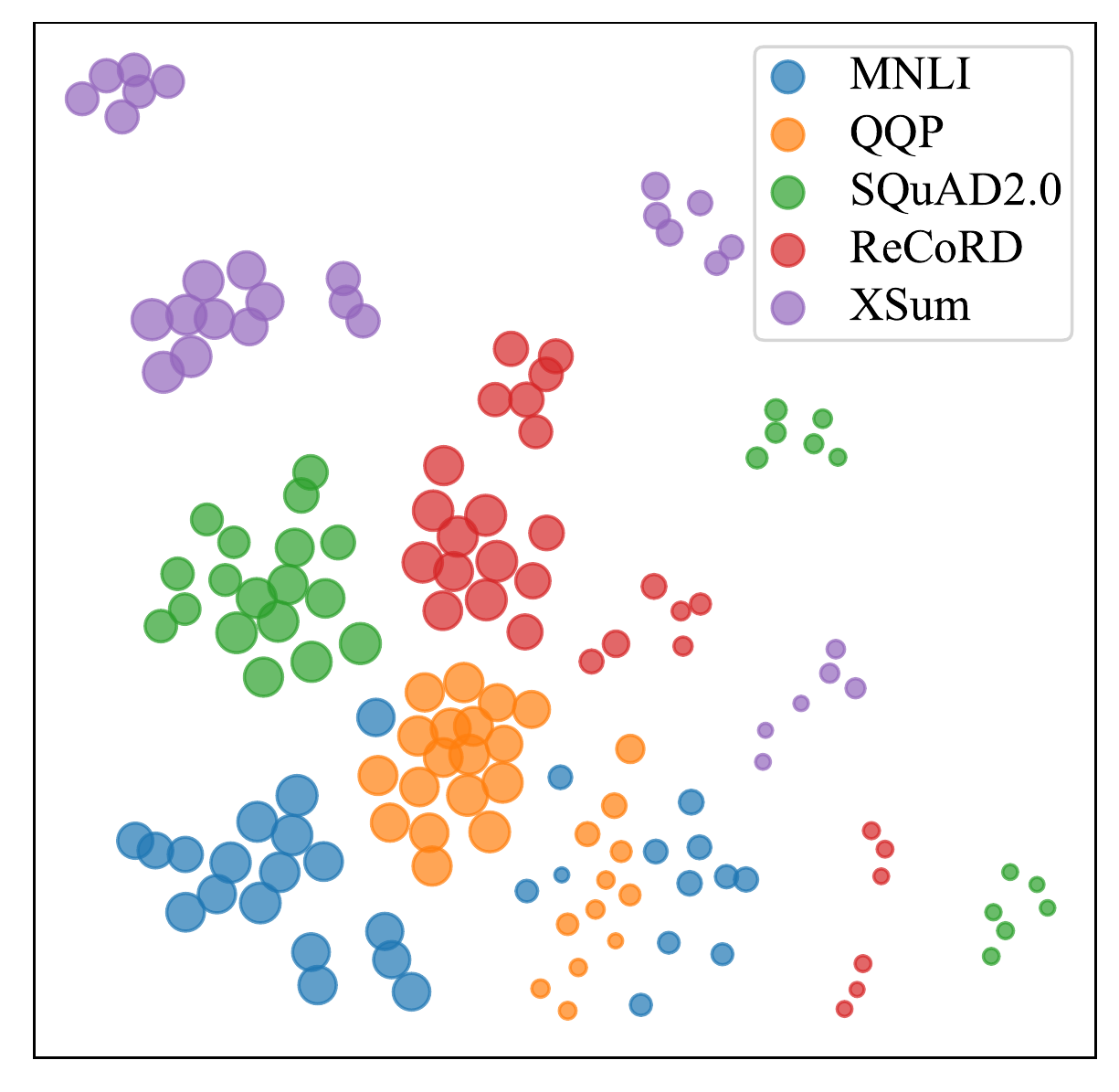} 
    \caption{Visualization of $5$ investigated tasks' soft prompts of different partial PLMs. A marker with a larger size means the performance of the corresponding soft prompts on the partial PLM is better.} 
    \label{fig:prompt_cluster} 
\end{figure}

\paragraph{Prompt Embedding Visualization.}
Taking a step further, we visualize the learned prompt embeddings of different partial PLMs using t-SNE~\citep{van2008visualizing} in Figure~\ref{fig:prompt_cluster}, and describe the details in \cref{sec:prompt_embedding_visulization}. We observe that \textbf{for the same task, the soft prompts obtained by different partial PLMs tend to form a compact cluster in the parameter space}. This phenomenon implies that the soft prompts corresponding to the same task (1) have a great potential of transferring among different partial PLMs, and (2) could serve as a better initialization that leads to faster convergence.
Apart from the visualization, we further report the cosine similarity of the learned prompts in \cref{sec:prompt_embedding_similarity} to verify the above phenomenon from another aspect.

\section{Fast Prompt Tuning}
In this section, we propose Fast Prompt Tuning (FPT), which aims at accelerating PT via \textit{progressive training}~\citep{pmlr-v97-gong19a}. Progressive training is typically leveraged for improving pre-training efficiency~\citep{chen-etal-2022-bert2bert,qin-etal-2022-elle}, instead, we focus on its application in PLM's downstream adaptation.
\subsection{Methodology}
Formally speaking, as visualized in Figure~\ref{fig:model} (c), we split the original PT training process into N stages. We start with a small-size partial PLM $\mathcal{M}_1$ and then progressively rehabilitate its depth and width until the full-size model $\mathcal{M}_\text{N}$, creating a series of partial PLMs $\{\mathcal{M}_i\}_{i=1}^{\text{N}-1}$ with growing sizes. The architectures of the partial PLMs are constructed using the same method in \cref{subsec:partial_model_construction}. 

During each training stage $i$, we conduct PT on a partial PLM $\mathcal{M}_i$ and obtain the learned soft prompts $\mathbf{P}_{i}$. Based on the observation that $\mathcal{M}_i$ retains a large portion of functionalities of the full-size PLM $\mathcal{M}_\text{N}$, we conjecture that $\mathcal{M}_i$ could serve as a perfect substitute for $\mathcal{M}_\text{N}$ and learn how to deal with the downstream task. In addition, considering that the soft prompts learned by different partial PLMs are close in the parameter space, we could transfer the knowledge learned by $\mathcal{M}_i$ to $\mathcal{M}_{i+1}$ through recycling $\mathbf{P}_{i}$. Specifically, after each model expansion, we directly use $\mathbf{P}_{i}$ as initialization for training $\mathcal{M}_{i+1}$ in the next stage. Since for each partial PLM, fewer parameters participate in both the forward and backward process, the computations could be reduced. Keeping the total number of training steps the same, FPT could accelerate training compared with vanilla PT.

\subsection{Experiments and Analyses}
\label{subsec:fpt_experiments}

We follow most of the experimental settings in \cref{sec:preliminary} and also describe the training details in \cref{sec:training_details}. We report FLOPs and training wall clock for the vanilla PT and \ourmodel to compare training efficiency. We evaluate both T5$_\texttt{LARGE}$ and T5$_\texttt{XL}$ (a larger T5 model) on each task and train for $30$k and $15$k steps, respectively. We test \ourmodel's performance when we progressively expand the model's depth, width, and both of them. Unless otherwise specified, for most of \ourmodel's methods, we split the training process into $4$ stages. Each of the first three stages takes $20$\% steps, while the last stage takes $40$\% steps.

\paragraph{Results.}
We list the results in Table~\ref{tab:results}, from which we observe that
(1) on average, all three variants of FPT achieve comparable performance with PT and utilize fewer computations (e.g., FPT$_{\text{CR}}$ saves around $30\%$ FLOPs). On several tasks (e.g., \textsc{MNLI} and \textsc{SQuAD2.0}), FPT even exceeds PT's performance;
(2) combining both layer dropping and FFN reduction (i.e., FPT$_{\text{CR}}$) is more training-efficient. However, we also observe that saving more computations generally leads to poorer performance. Among all three variants of FPT, $\text{FPT}_{\text{FR}}$ strikes the best balance between performance and training efficiency;
(3) moreover, we compare both PT and \ourmodel's performance when PT consumes the same computations as each variant of FPT. As reflected in the column ``Improve$\uparrow$'', controlling the training computations the same, our FPT outperforms PT, and the improvement is more significant for T5$_\texttt{XL}$ than T5$_\texttt{LARGE}$, showing that \ourmodel has a great potential to apply to large-scale PLMs.
(4) except for using FLOPs as a theoretical analysis of computation resources, we also compare wall clock training time among different FPT methods and vanilla PT. The wall clock time can be also saved at most 30\% with our $\text{FPT}_{\text{CR}}$ method. Besides, the gap between relative FLOPs and relative wall clock time shrinks with the model's size increasing for each FPT method.

We also verify the effectiveness of our partial model construction designs in \cref{subsec:effect_of_selction}, and show in \cref{subsec:effect_steps} that the performance of FPT is not sensitive to the duration of each training stage. We leave the explorations on other tasks and the effect of training budgets as future work.

\section{Conclusion}
In this work, towards improving PT's training efficiency, we first make several insightful observations by conducting PT on partial PLMs, and then propose \ourmodel based on the observations. The results on $5$ datasets demonstrate the feasibility of \ourmodel in saving the training computations. Being the first attempt towards accelerating PT, we encourage future work to design more sophisticated algorithms to further improve PT's training efficiency.

\section*{Acknowledgements}
This work is supported by the National Key R\&D Program of China (No. 2020AAA0106502) and Institute Guo Qiang at Tsinghua University.

Yufei Huang and Yujia Qin designed the methods, conducted the experiments, and wrote the paper. Huadong Wang and Yichun Yin participated in the discussion and provided valuable suggestions. Maosong Sun, Zhiyuan Liu, and Qun Liu advised the project.
\section*{Limitations}
For the current \ourmodel method, there exist two main limitations:

(1) \ourmodel requires choosing a proper hyper-parameter of the progressive training steps (i.e. duration of each training stage). For each experiment, we have to pre-define the duration of each stage empirically. Although in \cref{subsec:effect_steps}, we have shown that within a reasonable range, the duration of each training stage is not that important.

(2) \ourmodel can not be directly applied to other delta tuning methods (e.g., adapter and prefix-tuning). Since prompt tuning only adds trainable parameters in the embedding layer, when partial model's size increases, the trained soft prompt can be directly transferred to a larger partial model without any modification. But for other popular delta tuning methods, when the layer of partial model increases, we have to add newly initialized parameters. 

\bibliography{anthology,custom}

\begin{thebibliography}{38}
\expandafter\ifx\csname natexlab\endcsname\relax\def\natexlab#1{#1}\fi

\bibitem[{Ben~Zaken et~al.(2022)Ben~Zaken, Goldberg, and
  Ravfogel}]{zaken2021bitfit}
Elad Ben~Zaken, Yoav Goldberg, and Shauli Ravfogel. 2022.
\newblock \href {https://doi.org/10.18653/v1/2022.acl-short.1} {{B}it{F}it:
  Simple parameter-efficient fine-tuning for transformer-based masked
  language-models}.
\newblock In \emph{Proceedings of the 60th Annual Meeting of the Association
  for Computational Linguistics (Volume 2: Short Papers)}, pages 1--9, Dublin,
  Ireland. Association for Computational Linguistics.

\bibitem[{Brown et~al.(2020)Brown, Mann, Ryder, Subbiah, Kaplan, Dhariwal,
  Neelakantan, Shyam, Sastry, Askell, Agarwal, Herbert{-}Voss, Krueger,
  Henighan, Child, Ramesh, Ziegler, Wu, Winter, Hesse, Chen, Sigler, Litwin,
  Gray, Chess, Clark, Berner, McCandlish, Radford, Sutskever, and
  Amodei}]{gpt3}
Tom~B. Brown, Benjamin Mann, Nick Ryder, Melanie Subbiah, Jared Kaplan,
  Prafulla Dhariwal, Arvind Neelakantan, Pranav Shyam, Girish Sastry, Amanda
  Askell, Sandhini Agarwal, Ariel Herbert{-}Voss, Gretchen Krueger, Tom
  Henighan, Rewon Child, Aditya Ramesh, Daniel~M. Ziegler, Jeffrey Wu, Clemens
  Winter, Christopher Hesse, Mark Chen, Eric Sigler, Mateusz Litwin, Scott
  Gray, Benjamin Chess, Jack Clark, Christopher Berner, Sam McCandlish, Alec
  Radford, Ilya Sutskever, and Dario Amodei. 2020.
\newblock \href
  {https://proceedings.neurips.cc/paper/2020/hash/1457c0d6bfcb4967418bfb8ac142f64a-Abstract.html}
  {Language models are few-shot learners}.
\newblock In \emph{Advances in Neural Information Processing Systems 33: Annual
  Conference on Neural Information Processing Systems 2020, NeurIPS 2020,
  December 6-12, 2020, virtual}.

\bibitem[{Chen et~al.(2022)Chen, Yin, Shang, Jiang, Qin, Wang, Wang, Chen, Liu,
  and Liu}]{chen-etal-2022-bert2bert}
Cheng Chen, Yichun Yin, Lifeng Shang, Xin Jiang, Yujia Qin, Fengyu Wang, Zhi
  Wang, Xiao Chen, Zhiyuan Liu, and Qun Liu. 2022.
\newblock \href {https://doi.org/10.18653/v1/2022.acl-long.151} {bert2{BERT}:
  Towards reusable pretrained language models}.
\newblock In \emph{Proceedings of the 60th Annual Meeting of the Association
  for Computational Linguistics (Volume 1: Long Papers)}, pages 2134--2148,
  Dublin, Ireland. Association for Computational Linguistics.

\bibitem[{Chen et~al.(2020)Chen, Frankle, Chang, Liu, Zhang, Wang, and
  Carbin}]{chen2020lottery}
Tianlong Chen, Jonathan Frankle, Shiyu Chang, Sijia Liu, Yang Zhang, Zhangyang
  Wang, and Michael Carbin. 2020.
\newblock \href
  {https://proceedings.neurips.cc/paper/2020/hash/b6af2c9703f203a2794be03d443af2e3-Abstract.html}
  {The lottery ticket hypothesis for pre-trained {BERT} networks}.
\newblock In \emph{Advances in Neural Information Processing Systems 33: Annual
  Conference on Neural Information Processing Systems 2020, NeurIPS 2020,
  December 6-12, 2020, virtual}.

\bibitem[{Clark et~al.(2019)Clark, Khandelwal, Levy, and
  Manning}]{clark-etal-2019-bert}
Kevin Clark, Urvashi Khandelwal, Omer Levy, and Christopher~D. Manning. 2019.
\newblock \href {https://doi.org/10.18653/v1/W19-4828} {What does {BERT} look
  at? an analysis of {BERT}{'}s attention}.
\newblock In \emph{Proceedings of the 2019 ACL Workshop BlackboxNLP: Analyzing
  and Interpreting Neural Networks for NLP}, pages 276--286, Florence, Italy.
  Association for Computational Linguistics.

\bibitem[{Devlin et~al.(2019)Devlin, Chang, Lee, and
  Toutanova}]{devlin2018bert}
Jacob Devlin, Ming-Wei Chang, Kenton Lee, and Kristina Toutanova. 2019.
\newblock \href {https://doi.org/10.18653/v1/N19-1423} {{BERT}: Pre-training of
  deep bidirectional transformers for language understanding}.
\newblock In \emph{Proceedings of the 2019 Conference of the North {A}merican
  Chapter of the Association for Computational Linguistics: Human Language
  Technologies, Volume 1 (Long and Short Papers)}, pages 4171--4186,
  Minneapolis, Minnesota. Association for Computational Linguistics.

\bibitem[{Ding et~al.(2022)Ding, Qin, Yang, Wei, Yang, Su, Hu, Chen, Chan, Chen
  et~al.}]{ding2022delta}
Ning Ding, Yujia Qin, Guang Yang, Fuchao Wei, Zonghan Yang, Yusheng Su,
  Shengding Hu, Yulin Chen, Chi-Min Chan, Weize Chen, et~al. 2022.
\newblock \href {https://arxiv.org/pdf/2203.06904.pdf} {Delta tuning: A
  comprehensive study of parameter efficient methods for pre-trained language
  models}.
\newblock \emph{arXiv preprint arXiv:2203.06904}.

\bibitem[{Fan et~al.(2020)Fan, Grave, and Joulin}]{Fan2020Reducing}
Angela Fan, Edouard Grave, and Armand Joulin. 2020.
\newblock \href {https://openreview.net/forum?id=SylO2yStDr} {Reducing
  transformer depth on demand with structured dropout}.
\newblock In \emph{8th International Conference on Learning Representations,
  {ICLR} 2020, Addis Ababa, Ethiopia, April 26-30, 2020}. OpenReview.net.

\bibitem[{Gong et~al.(2019)Gong, He, Li, Qin, Wang, and Liu}]{pmlr-v97-gong19a}
Linyuan Gong, Di~He, Zhuohan Li, Tao Qin, Liwei Wang, and Tie{-}Yan Liu. 2019.
\newblock \href {http://proceedings.mlr.press/v97/gong19a.html} {Efficient
  training of {BERT} by progressively stacking}.
\newblock In \emph{Proceedings of the 36th International Conference on Machine
  Learning, {ICML} 2019, 9-15 June 2019, Long Beach, California, {USA}},
  volume~97 of \emph{Proceedings of Machine Learning Research}, pages
  2337--2346. {PMLR}.

\bibitem[{Gu et~al.(2021)Gu, Liu, Yu, Li, Chen, and Han}]{gu2020transformer}
Xiaotao Gu, Liyuan Liu, Hongkun Yu, Jing Li, Chen Chen, and Jiawei Han. 2021.
\newblock \href {https://doi.org/10.18653/v1/2021.naacl-main.406} {On the
  transformer growth for progressive {BERT} training}.
\newblock In \emph{Proceedings of the 2021 Conference of the North American
  Chapter of the Association for Computational Linguistics: Human Language
  Technologies}, pages 5174--5180, Online. Association for Computational
  Linguistics.

\bibitem[{Han et~al.(2021)Han, Zhang, Ding, Gu, Liu, Huo, Qiu, Yao, Zhang,
  Zhang et~al.}]{han2021pre}
Xu~Han, Zhengyan Zhang, Ning Ding, Yuxian Gu, Xiao Liu, Yuqi Huo, Jiezhong Qiu,
  Yuan Yao, Ao~Zhang, Liang Zhang, et~al. 2021.
\newblock \href {https://arxiv.org/pdf/2106.07139.pdf} {Pre-trained models:
  Past, present and future}.
\newblock \emph{AI Open}, 2:225--250.

\bibitem[{He et~al.(2022)He, Zhou, Ma, Berg{-}Kirkpatrick, and
  Neubig}]{he2021towards}
Junxian He, Chunting Zhou, Xuezhe Ma, Taylor Berg{-}Kirkpatrick, and Graham
  Neubig. 2022.
\newblock \href {https://openreview.net/forum?id=0RDcd5Axok} {Towards a unified
  view of parameter-efficient transfer learning}.
\newblock In \emph{The Tenth International Conference on Learning
  Representations, {ICLR} 2022, Virtual Event, April 25-29, 2022}.
  OpenReview.net.

\bibitem[{Houlsby et~al.(2019)Houlsby, Giurgiu, Jastrzebski, Morrone,
  de~Laroussilhe, Gesmundo, Attariyan, and Gelly}]{houlsby2019parameter}
Neil Houlsby, Andrei Giurgiu, Stanislaw Jastrzebski, Bruna Morrone, Quentin
  de~Laroussilhe, Andrea Gesmundo, Mona Attariyan, and Sylvain Gelly. 2019.
\newblock \href {http://proceedings.mlr.press/v97/houlsby19a.html}
  {Parameter-efficient transfer learning for {NLP}}.
\newblock In \emph{Proceedings of the 36th International Conference on Machine
  Learning, {ICML} 2019, 9-15 June 2019, Long Beach, California, {USA}},
  volume~97 of \emph{Proceedings of Machine Learning Research}, pages
  2790--2799. {PMLR}.

\bibitem[{Hu et~al.(2022)Hu, Shen, Wallis, Allen{-}Zhu, Li, Wang, Wang, and
  Chen}]{hu2021lora}
Edward~J. Hu, Yelong Shen, Phillip Wallis, Zeyuan Allen{-}Zhu, Yuanzhi Li,
  Shean Wang, Lu~Wang, and Weizhu Chen. 2022.
\newblock \href {https://openreview.net/forum?id=nZeVKeeFYf9} {Lora: Low-rank
  adaptation of large language models}.
\newblock In \emph{The Tenth International Conference on Learning
  Representations, {ICLR} 2022, Virtual Event, April 25-29, 2022}.
  OpenReview.net.

\bibitem[{Jaszczur et~al.(2021)Jaszczur, Chowdhery, Mohiuddin, Kaiser,
  Gajewski, Michalewski, and Kanerva}]{jaszczur2021sparse}
Sebastian Jaszczur, Aakanksha Chowdhery, Afroz Mohiuddin, Lukasz Kaiser,
  Wojciech Gajewski, Henryk Michalewski, and Jonni Kanerva. 2021.
\newblock \href
  {https://proceedings.neurips.cc/paper/2021/hash/51f15efdd170e6043fa02a74882f0470-Abstract.html}
  {Sparse is enough in scaling transformers}.
\newblock In \emph{Advances in Neural Information Processing Systems 34: Annual
  Conference on Neural Information Processing Systems 2021, NeurIPS 2021,
  December 6-14, 2021, virtual}, pages 9895--9907.

\bibitem[{Jawahar et~al.(2019)Jawahar, Sagot, and
  Seddah}]{jawahar-etal-2019-bert}
Ganesh Jawahar, Beno{\^\i}t Sagot, and Djam{\'e} Seddah. 2019.
\newblock \href {https://doi.org/10.18653/v1/P19-1356} {What does {BERT} learn
  about the structure of language?}
\newblock In \emph{Proceedings of the 57th Annual Meeting of the Association
  for Computational Linguistics}, pages 3651--3657, Florence, Italy.
  Association for Computational Linguistics.

\bibitem[{Lester et~al.(2021)Lester, Al-Rfou, and
  Constant}]{lester-etal-2021-power}
Brian Lester, Rami Al-Rfou, and Noah Constant. 2021.
\newblock \href {https://aclanthology.org/2021.emnlp-main.243} {The power of
  scale for parameter-efficient prompt tuning}.
\newblock In \emph{Proceedings of the 2021 Conference on Empirical Methods in
  Natural Language Processing}, pages 3045--3059, Online and Punta Cana,
  Dominican Republic. Association for Computational Linguistics.

\bibitem[{Li and Liang(2021)}]{li-liang-2021-prefix}
Xiang~Lisa Li and Percy Liang. 2021.
\newblock \href {https://doi.org/10.18653/v1/2021.acl-long.353} {Prefix-tuning:
  Optimizing continuous prompts for generation}.
\newblock In \emph{Proceedings of the 59th Annual Meeting of the Association
  for Computational Linguistics and the 11th International Joint Conference on
  Natural Language Processing (Volume 1: Long Papers)}, pages 4582--4597,
  Online. Association for Computational Linguistics.

\bibitem[{Liu et~al.(2019)Liu, Ott, Goyal, Du, Joshi, Chen, Levy, Lewis,
  Zettlemoyer, and Stoyanov}]{liu2019roberta}
Yinhan Liu, Myle Ott, Naman Goyal, Jingfei Du, Mandar Joshi, Danqi Chen, Omer
  Levy, Mike Lewis, Luke Zettlemoyer, and Veselin Stoyanov. 2019.
\newblock \href {http://arxiv.org/abs/1907.11692} {Ro{BERT}a: A robustly
  optimized {BERT} pretraining approach}.

\bibitem[{Narayan et~al.(2018)Narayan, Cohen, and
  Lapata}]{narayan-etal-2018-dont}
Shashi Narayan, Shay~B. Cohen, and Mirella Lapata. 2018.
\newblock \href {https://doi.org/10.18653/v1/D18-1206} {Don{'}t give me the
  details, just the summary! topic-aware convolutional neural networks for
  extreme summarization}.
\newblock In \emph{Proceedings of the 2018 Conference on Empirical Methods in
  Natural Language Processing}, pages 1797--1807, Brussels, Belgium.
  Association for Computational Linguistics.

\bibitem[{Paszke et~al.(2019)Paszke, Gross, Massa, Lerer, Bradbury, Chanan,
  Killeen, Lin, Gimelshein, Antiga, Desmaison, K{\"{o}}pf, Yang, DeVito,
  Raison, Tejani, Chilamkurthy, Steiner, Fang, Bai, and
  Chintala}]{DBLP:conf/nips/PaszkeGMLBCKLGA19}
Adam Paszke, Sam Gross, Francisco Massa, Adam Lerer, James Bradbury, Gregory
  Chanan, Trevor Killeen, Zeming Lin, Natalia Gimelshein, Luca Antiga, Alban
  Desmaison, Andreas K{\"{o}}pf, Edward~Z. Yang, Zachary DeVito, Martin Raison,
  Alykhan Tejani, Sasank Chilamkurthy, Benoit Steiner, Lu~Fang, Junjie Bai, and
  Soumith Chintala. 2019.
\newblock \href
  {https://proceedings.neurips.cc/paper/2019/hash/bdbca288fee7f92f2bfa9f7012727740-Abstract.html}
  {Pytorch: An imperative style, high-performance deep learning library}.
\newblock In \emph{Advances in Neural Information Processing Systems 32: Annual
  Conference on Neural Information Processing Systems 2019, NeurIPS 2019,
  December 8-14, 2019, Vancouver, BC, Canada}, pages 8024--8035.

\bibitem[{Qin et~al.(2021)Qin, Wang, Su, Lin, Ding, Liu, Li, Hou, Li, Sun
  et~al.}]{qin2021exploring}
Yujia Qin, Xiaozhi Wang, Yusheng Su, Yankai Lin, Ning Ding, Zhiyuan Liu, Juanzi
  Li, Lei Hou, Peng Li, Maosong Sun, et~al. 2021.
\newblock \href {https://arxiv.org/pdf/2110.07867.pdf} {Exploring
  low-dimensional intrinsic task subspace via prompt tuning}.
\newblock \emph{arXiv preprint arXiv:2110.07867}.

\bibitem[{Qin et~al.(2022)Qin, Zhang, Lin, Liu, Li, Sun, and
  Zhou}]{qin-etal-2022-elle}
Yujia Qin, Jiajie Zhang, Yankai Lin, Zhiyuan Liu, Peng Li, Maosong Sun, and Jie
  Zhou. 2022.
\newblock \href {https://doi.org/10.18653/v1/2022.findings-acl.220} {{ELLE}:
  Efficient lifelong pre-training for emerging data}.
\newblock In \emph{Findings of the Association for Computational Linguistics:
  ACL 2022}, pages 2789--2810, Dublin, Ireland. Association for Computational
  Linguistics.

\bibitem[{Raffel et~al.(2020)Raffel, Shazeer, Roberts, Lee, Narang, Matena,
  Zhou, Li, and Liu}]{2020t5}
Colin Raffel, Noam Shazeer, Adam Roberts, Katherine Lee, Sharan Narang, Michael
  Matena, Yanqi Zhou, Wei Li, and Peter~J. Liu. 2020.
\newblock \href {http://jmlr.org/papers/v21/20-074.html} {Exploring the limits
  of transfer learning with a unified text-to-text transformer}.
\newblock \emph{Journal of Machine Learning Research}, 21(140):1--67.

\bibitem[{Rajpurkar et~al.(2018)Rajpurkar, Jia, and
  Liang}]{rajpurkar-etal-2018-know}
Pranav Rajpurkar, Robin Jia, and Percy Liang. 2018.
\newblock \href {https://doi.org/10.18653/v1/P18-2124} {Know what you don{'}t
  know: Unanswerable questions for {SQ}u{AD}}.
\newblock In \emph{Proceedings of the 56th Annual Meeting of the Association
  for Computational Linguistics (Volume 2: Short Papers)}, pages 784--789,
  Melbourne, Australia. Association for Computational Linguistics.

\bibitem[{Shazeer and Stern(2018)}]{pmlr-v80-shazeer18a}
Noam Shazeer and Mitchell Stern. 2018.
\newblock \href {http://proceedings.mlr.press/v80/shazeer18a.html} {Adafactor:
  Adaptive learning rates with sublinear memory cost}.
\newblock In \emph{Proceedings of the 35th International Conference on Machine
  Learning, {ICML} 2018, Stockholmsm{\"{a}}ssan, Stockholm, Sweden, July 10-15,
  2018}, volume~80 of \emph{Proceedings of Machine Learning Research}, pages
  4603--4611. {PMLR}.

\bibitem[{Su et~al.(2022)Su, Wang, Qin, Chan, Lin, Wang, Wen, Liu, Li, Li, Hou,
  Sun, and Zhou}]{su2021transferability}
Yusheng Su, Xiaozhi Wang, Yujia Qin, Chi-Min Chan, Yankai Lin, Huadong Wang,
  Kaiyue Wen, Zhiyuan Liu, Peng Li, Juanzi Li, Lei Hou, Maosong Sun, and Jie
  Zhou. 2022.
\newblock \href {https://doi.org/10.18653/v1/2022.naacl-main.290} {On
  transferability of prompt tuning for natural language processing}.
\newblock In \emph{Proceedings of the 2022 Conference of the North American
  Chapter of the Association for Computational Linguistics: Human Language
  Technologies}, pages 3949--3969, Seattle, United States. Association for
  Computational Linguistics.

\bibitem[{Sun et~al.(2020)Sun, Yu, Song, Liu, Yang, and
  Zhou}]{sun-etal-2020-mobilebert}
Zhiqing Sun, Hongkun Yu, Xiaodan Song, Renjie Liu, Yiming Yang, and Denny Zhou.
  2020.
\newblock \href {https://doi.org/10.18653/v1/2020.acl-main.195}
  {{M}obile{BERT}: a compact task-agnostic {BERT} for resource-limited
  devices}.
\newblock In \emph{Proceedings of the 58th Annual Meeting of the Association
  for Computational Linguistics}, pages 2158--2170, Online. Association for
  Computational Linguistics.

\bibitem[{Teerapittayanon et~al.(2016)Teerapittayanon, McDanel, and
  Kung}]{teerapittayanon2016branchynet}
Surat Teerapittayanon, Bradley McDanel, and Hsiang-Tsung Kung. 2016.
\newblock \href {https://arxiv.org/pdf/1709.01686.pdf} {Branchynet: Fast
  inference via early exiting from deep neural networks}.
\newblock In \emph{2016 23rd International Conference on Pattern Recognition
  (ICPR)}, pages 2464--2469. IEEE.

\bibitem[{van~der Maaten and Hinton(2008)}]{van2008visualizing}
Laurens van~der Maaten and Geoffrey Hinton. 2008.
\newblock \href {http://jmlr.org/papers/v9/vandermaaten08a.html} {Visualizing
  data using t-sne}.
\newblock \emph{Journal of Machine Learning Research}, 9(86):2579--2605.

\bibitem[{Vaswani et~al.(2017)Vaswani, Shazeer, Parmar, Uszkoreit, Jones,
  Gomez, Kaiser, and Polosukhin}]{vaswani2017attention}
Ashish Vaswani, Noam Shazeer, Niki Parmar, Jakob Uszkoreit, Llion Jones,
  Aidan~N. Gomez, Lukasz Kaiser, and Illia Polosukhin. 2017.
\newblock \href
  {https://proceedings.neurips.cc/paper/2017/hash/3f5ee243547dee91fbd053c1c4a845aa-Abstract.html}
  {Attention is all you need}.
\newblock In \emph{Advances in Neural Information Processing Systems 30: Annual
  Conference on Neural Information Processing Systems 2017, December 4-9, 2017,
  Long Beach, CA, {USA}}, pages 5998--6008.

\bibitem[{Vu et~al.(2022)Vu, Lester, Constant, Al-Rfou{'}, and
  Cer}]{vu-etal-2022-spot}
Tu~Vu, Brian Lester, Noah Constant, Rami Al-Rfou{'}, and Daniel Cer. 2022.
\newblock \href {https://doi.org/10.18653/v1/2022.acl-long.346} {{SP}o{T}:
  Better frozen model adaptation through soft prompt transfer}.
\newblock In \emph{Proceedings of the 60th Annual Meeting of the Association
  for Computational Linguistics (Volume 1: Long Papers)}, pages 5039--5059,
  Dublin, Ireland. Association for Computational Linguistics.

\bibitem[{Williams et~al.(2018)Williams, Nangia, and
  Bowman}]{williams-etal-2018-broad}
Adina Williams, Nikita Nangia, and Samuel Bowman. 2018.
\newblock \href {https://doi.org/10.18653/v1/N18-1101} {A broad-coverage
  challenge corpus for sentence understanding through inference}.
\newblock In \emph{Proceedings of the 2018 Conference of the North {A}merican
  Chapter of the Association for Computational Linguistics: Human Language
  Technologies, Volume 1 (Long Papers)}, pages 1112--1122, New Orleans,
  Louisiana. Association for Computational Linguistics.

\bibitem[{Wolf et~al.(2020)Wolf, Debut, Sanh, Chaumond, Delangue, Moi, Cistac,
  Rault, Louf, Funtowicz, Davison, Shleifer, von Platen, Ma, Jernite, Plu, Xu,
  Scao, Gugger, Drame, Lhoest, and Rush}]{wolf-etal-2020-transformers}
Thomas Wolf, Lysandre Debut, Victor Sanh, Julien Chaumond, Clement Delangue,
  Anthony Moi, Pierric Cistac, Tim Rault, Rémi Louf, Morgan Funtowicz, Joe
  Davison, Sam Shleifer, Patrick von Platen, Clara Ma, Yacine Jernite, Julien
  Plu, Canwen Xu, Teven~Le Scao, Sylvain Gugger, Mariama Drame, Quentin Lhoest,
  and Alexander~M. Rush. 2020.
\newblock \href {https://www.aclweb.org/anthology/2020.emnlp-demos.6}
  {Transformers: State-of-the-art natural language processing}.
\newblock In \emph{Proceedings of the 2020 Conference on Empirical Methods in
  Natural Language Processing: System Demonstrations}, pages 38--45, Online.
  Association for Computational Linguistics.

\bibitem[{Xin et~al.(2020)Xin, Tang, Lee, Yu, and Lin}]{xin2020deebert}
Ji~Xin, Raphael Tang, Jaejun Lee, Yaoliang Yu, and Jimmy Lin. 2020.
\newblock \href {https://doi.org/10.18653/v1/2020.acl-main.204} {{D}ee{BERT}:
  Dynamic early exiting for accelerating {BERT} inference}.
\newblock In \emph{Proceedings of the 58th Annual Meeting of the Association
  for Computational Linguistics}, pages 2246--2251, Online. Association for
  Computational Linguistics.

\bibitem[{Zhang and He(2020)}]{NEURIPS2020_a1140a3d}
Minjia Zhang and Yuxiong He. 2020.
\newblock \href
  {https://proceedings.neurips.cc/paper/2020/file/a1140a3d0df1c81e24ae954d935e8926-Paper.pdf}
  {Accelerating training of transformer-based language models with progressive
  layer dropping}.
\newblock In \emph{Advances in Neural Information Processing Systems},
  volume~33, pages 14011--14023. Curran Associates, Inc.

\bibitem[{Zhang et~al.(2018)Zhang, Liu, Liu, Gao, Duh, and
  Durme}]{zhang2018record}
Sheng Zhang, Xiaodong Liu, Jingjing Liu, Jianfeng Gao, Kevin Duh, and
  Benjamin~Van Durme. 2018.
\newblock \href {http://arxiv.org/abs/1810.12885} {Re{C}o{RD}: Bridging the gap
  between human and machine commonsense reading comprehension}.

\bibitem[{Zhang et~al.(2022)Zhang, Lin, Liu, Li, Sun, and
  Zhou}]{zhang2021moefication}
Zhengyan Zhang, Yankai Lin, Zhiyuan Liu, Peng Li, Maosong Sun, and Jie Zhou.
  2022.
\newblock \href {https://doi.org/10.18653/v1/2022.findings-acl.71}
  {{M}o{E}fication: Transformer feed-forward layers are mixtures of experts}.
\newblock In \emph{Findings of the Association for Computational Linguistics:
  ACL 2022}, pages 877--890, Dublin, Ireland. Association for Computational
  Linguistics.

\end{thebibliography}
\bibliographystyle{acl_natbib}

\clearpage
\appendix

\section*{Appendices}
\label{sec:appendix}

\section{Related Work}
\paragraph{Prompt Tuning.} PLMs have achieved excellent performance on many NLP tasks relying on their powerful natural language understanding and generation capabilities~\citep{devlin2018bert, liu2019roberta}. However, with the emergence of large-scale PLMs like T5~\citep{2020t5} and GPT-3~\citep{gpt3}, tuning all the parameters of a PLM (i.e., full-parameter fine-tuning), which requires huge storage and memory costs, is not flexible for real-world applications on massive downstream tasks. Therefore, parameter-efficient delta tuning methods \citep{ding2022delta,houlsby2019parameter,hu2021lora,zaken2021bitfit,he2021towards} attract more and more attention, among which prompt tuning (PT) \citep{lester-etal-2021-power} is a simple and effective one. By prepending a few trainable embeddings before the input sequence, PT can achieve comparable performance to full-parameter fine-tuning. With the size of PLM getting larger, the performance of PT gets closer to vanilla fine-tuning~\citep{lester-etal-2021-power}, showing great potential to utilize extremely large PLMs in future. Besides, PT is also shown to have excellent cross-task transferability~\citep{su2021transferability,vu-etal-2022-spot}, and thus gains more and more attention in exploring the relation among tasks~\citep{qin2021exploring}. However, due to the slow convergence shown in Figure~\ref{fig:finetuning_prompttuning}, PT's training efficiency becomes a serious drawback and may limit its practical application.

\paragraph{Progressive Training.} Considering that pre-training usually requires tremendous computational resources, researchers propose \textit{progressive training} to improve training efficiency \citep{pmlr-v97-gong19a, NEURIPS2020_a1140a3d}. Progressive training starts training using a shallow model, and gradually grows the depth of the model along the training process by replicating existing layers (parameter recycling). In this way, the pre-training efficiency can be improved a lot. To further improve training efficiency, later works propose to progressively grow PLMs in both depth and width~\citep{gu2020transformer}, and design better initialization methods to inherit the functionality of existing models~\citep{chen-etal-2022-bert2bert}. Instead of leveraging progressive training during the process of pre-training, we apply it to PLM's downstream adaptation, with a focus on PT. Furthermore, conventional progressive training duplicates existing parameters to grow a PLM's size until the full-model's size. Instead, we have already obtained a full-size PLM, and propose to construct partial models with growing sizes by dropping / masking existing parameters.

\section{Implementation Details}
\label{sec:training_details}
Our implementation is based on PyTorch~\citep{DBLP:conf/nips/PaszkeGMLBCKLGA19} and transformers~\citep{wolf-etal-2020-transformers}. The experiments are conducted with $8$ NVIDIA 32GB V100 GPUs, and each experiment requires fewer than $10$ hours to finish.

\paragraph{Partial PLM Construction.} As mentioned in \cref{{subsec:partial_model_construction}}, we design three methods to construct partial PLMs. Specifically, for \textbf{layer dropping}, we select layers uniformly. For example, to select $3$ layers out of a $24$-layer PLM, we will select layer $\{1, 12, 24\}$ to construct the partial PLM. For \textbf{FFN reduction}, to estimate the activation of each neuron (dimension) in FFN layer $l$, we first randomly sample $1,000$ examples to form a small dataset $\mathcal{D}$. We prepend each example $\mathcal{X}$ (without the label) in $\mathcal{D}$ with randomly initialized soft prompts and feed it into the full-size PLM to obtain the input representation $\bm{x}^l$ of FFN layer. After that, we obtain the activation score of each neuron using the following equation $S =\sum_{\mathcal{X} \in \mathcal{D}} \sum_{i=1}^{|\mathcal{X}|}\left| \sigma(\bm{x}_i^l\bm{W}_1^l+\bm{b}_1^l) \right|$, where $\bm{W}_1^l, \bm{b}_1^l$ are the parameters in FFN layer $l$, and $|\mathcal{X}|$ denotes the sequence length. The neurons (dimensions) with smaller activation score (i.e., seldom activated) will be masked. Note that the T5 model is composed of both an encoder and a decoder, due to the difference in the input length and output length on various tasks, the computation overload of the encoder and decoder may vary a lot. Therefore, for the tasks ($\textsc{MNLI}$ and $\textsc{QQP}$) that have a lighter computation overload on the decoder (i.e., small output length), shrinking the decoder model size has little impact on saving the computational costs, hence we retain the whole decoder under this setting; for other three tasks ($\textsc{SQuAD2.0}$, $\textsc{ReCoRD}$ and $\textsc{XSum}$), the output length on decoder is much longer and we conduct partial PLM construction on both the encoder and decoder. We calculate FLOPs for each experiment using the \textit{ptflops} tool \footnote{\url{https://github.com/sovrasov/flops-counter.pytorch}}, and report the average FLOPs of $5$ tasks in Table~\ref{tab:preliminary_results} and Table~\ref{tab:results}.

\begin{table*}[!t]
    \centering
    \small
    \begin{tabular}{lc|cc|cc|cc|c}
        \toprule[1pt]
  \multicolumn{2}{l|}{\multirow{2}{*}{Partial PLMs}} & \multicolumn{2}{c|}{Layer Dropping} & \multicolumn{2}{c|}{FFN Reduction} & \multicolumn{2}{c|}{Compound Reduction} & \multirow{2}*{Training Steps} \\
        \multicolumn{2}{l|}{\ } & $L_{enc} / L_{dec}$ & $d'$ & $L_{enc} / L_{dec}$ & $d'$ & $L_{enc} / L_{dec}$ & $d'$ & \\
        \midrule[1pt]
        \multirow{4}*{T5$_{\texttt{LARGE}}$}
        & $\mathcal{M}_1$ & 6 / 6 & 2,816 & 24 / 24 & 704 & 6 / 6 & 704 & 6,000 \\
        & $\mathcal{M}_2$ & 12 / 12 & 2,816 & 24 / 24 & 1,408 & 12 / 12 & 1,408 & 6,000 \\
        & $\mathcal{M}_3$ & 18 / 18 & 2,816 & 24 / 24 & 2,112 & 18 / 18 & 2,112 & 6,000 \\
        & $\mathcal{M}_4$ & 24 / 24 & 2,816 & 24 / 24 & 2,816 & 24 / 24 & 2,816 & 12,000 \\
        \midrule[1pt]
        \multirow{4}*{T5$_{\texttt{XL}}$}
        & $\mathcal{M}_1$ & 18 / 18  & 5,120 & 24 / 24 & 1,280 & 18 / 18  & 1,280 & 3,000 \\
        & $\mathcal{M}_2$ & 18 / 18 & 5,120 & 24 / 24 & 2,560 & 18 / 18 & 2,560 & 3,000 \\
        & $\mathcal{M}_3$ & 18 / 18 & 5,120 & 24 / 24 & 3,840 & 18 / 18 & 3,840 & 3,000 \\
        & $\mathcal{M}_4$ & 24 / 24 & 5,120 & 24 / 24 & 5,120 & 24 / 24 & 5,120 & 6,000 \\
        \bottomrule[1pt]
    \end{tabular}
    \caption{Architecture details of the partial PLMs on the three construction methods for both $\text{T5}_{\texttt{LARGE}}$ and $\text{T5}_{\texttt{XL}}$. $L_{enc}$ and $L_{dec}$ denote the number of layers of the encoder and decoder of the partial model $\mathcal{M}_i$, respectively. We also list the training steps for each stage in the last column.}
    \label{tab:progressive_method}
\end{table*}

\paragraph{Partial PLM Prompt Tuning.}
We use T5$_{\texttt{LARGE}}$ for our experiments of partial PLM PT. Following \citet{lester-etal-2021-power}, we leverage the LM-adapted version of T5 checkpoints, which are additionally trained for $100$k steps. The adapted version of T5 has been demonstrated to achieve stable and better PT performance. For the implementation of PT, we set the prompt length to $20$ and randomly initialize the soft prompts. The optimizer is chosen as Adafactor~\citep{pmlr-v80-shazeer18a} and the learning rate is set to $0.3$. We choose a batch size of $32$ and greedy decoding to generate the predictions. The training steps are set to $30$k to ensure that PT will not get stuck in a local optimum. We run all the experiments $3$ times with different random seeds and report the average results.

\begin{table*}[!t]
    \centering
    \small
    \begin{tabular}{l|c|c|c|c|c}
    \toprule
        \diagbox{$\mathcal{T}^P$}{$\mathcal{T}^F$} & $\textsc{MNLI}$ & $\textsc{QQP}$ & $\textsc{SQuAD2.0}$ & $\textsc{ReCoRD}$ & $\textsc{XSum}$ \\
        \midrule
        $\textsc{MNLI}$ & \textbf{0.249} & 0.131 & 0.175 & 0.109 & 0.139 \\
        $\textsc{QQP}$ & 0.145 & \textbf{0.177} & 0.135 & 0.103 & 0.126 \\
        $\textsc{SQuAD2.0}$ & 0.202 & 0.143 & \textbf{0.286} & 0.190 & 0.202 \\
        $\textsc{ReCoRD}$ & 0.167 & 0.119 & 0.219 & 0.\textbf{224} & 0.195 \\
        $\textsc{XSum}$ & 0.164 & 0.128 & 0.237 & 0.188 & \textbf{0.301} \\
        \bottomrule
    \end{tabular}
    \caption{Average prompt similarity ($\mathcal{S}(\mathcal{T}_j^P, \mathcal{T}_k^F)$) among different tasks. The highest score in each row is \textbf{highlighted}.}
    \label{tab:similarity}
\end{table*}

\paragraph{Fast Prompt Tuning.}
For the implementations of \ourmodel, we train T5$_\texttt{LARGE}$ / T5$_\texttt{XL}$ with a total step of $30$k / $15$k. The number of training steps of T5$_\texttt{XL}$ is chosen smaller than T5$_\texttt{LARGE}$ since we find empirically that T5$_\texttt{XL}$ converges faster than T5$_\texttt{LARGE}$. As mentioned in \cref{subsec:fpt_experiments}, unless otherwise specified, we split the whole training process into $4$ stages. Each of the first three stages takes $20$\% of the training steps, while the last stage (full-model PT) takes $40$\% training stages. Except for layer dropping on T5$_{\texttt{XL}}$, we find that a partial PLM, with fewer than $12$ layers in either the encoder or decoder, achieves poor PT performance. Therefore, we only use two training stages where the first stage takes $60$\% training steps and the second stage takes $40$\% training steps. More detailed settings about the partial model construction are shown in Table~\ref{tab:progressive_method}. The experiments with T5$_{\texttt{LARGE}}$ are run three times with different random seeds and the average results are reported while experiments with T5$_{\texttt{XL}}$ are conducted once due to their huge computation consumption.

\section{Prompt Embedding Visualization}
\label{sec:prompt_embedding_visulization}

In Figure~\ref{fig:prompt_cluster}, we visualize the soft prompts of different partial PLMs and tasks in Table~\ref{tab:preliminary_results}. The embedding used for visualization is derived by averaging soft prompt along the token length dimension. As described in \cref{subsec:partial_model_experiment}, we run each experiment three times with different random seeds to get stable results. Therefore, we plot $30$ points ($3$ runs $\times$ ($9$ partial PLM + $1$ full-size PLM)) for each task in Figure~\ref{fig:prompt_cluster}. And the size of the marker in the figure denotes the performance of the soft prompts on corresponding partial PLMs. Larger size indicates better performance. We can observe that soft prompts with better performance will be easier to form a compact cluster.

\section{Prompt Embedding Similarity}
\label{sec:prompt_embedding_similarity}
To further gain insights on the transferability of the soft prompts learned by T5$_\texttt{LARGE}$'s different partial PLMs defined in Table~\ref{tab:progressive_method}, in addition to the visualization conducted in \cref{subsec:partial_model_experiment}, we calculate the average cosine similarity of the soft prompts corresponding to different tasks in Table~\ref{tab:similarity}. Specifically, for different partial PLMs $\mathcal{M}_1, \mathcal{M}_2, ...,\mathcal{M}_{\text{N}-1}$ and the full-size PLM $\mathcal{M}_{\text{N}}$, we conduct PT with each model $\mathcal{M}_i$ on the task $\mathcal{T}_j$ and obtain the corresponding soft prompts $\mathbf{P}_i^j \in \mathbb{R}^{l \times d}$. Then we average $\mathbf{P}_i^j$ along the token length dimension and get a vector $\overline{\mathbf{P}}^j_i \in \mathbb{R}^{d}$. After that, we calculate $\mathcal{S}(\mathcal{T}_j^P, \mathcal{T}_k^F)$ (average cosine similarity between (1) task $j$'s partial PLMs' prompts and (2) task $k$'s full-size PLM's prompts) using the following metric:

\begin{equation}
    \mathcal{S}(\mathcal{T}_j^P, \mathcal{T}_k^F) = \frac{1}{\text{N} - 1}\sum_{i = 1}^{\text{N}-1}\frac{\overline{\mathbf{P}}^j_i \cdot {\overline{\mathbf{P}}^k_\text{N}}}{\lVert \overline{\mathbf{P}}^j_i \rVert \lVert \overline{\mathbf{P}}^k_\text{N} \rVert}
\end{equation}

From the results in Table~\ref{tab:similarity}, we observe that the highest similarity is achieved when $j = k$, showing that the prompts of the partial PLMs are closer to the same task's prompts of the full-size model. This phenomenon is aligned with the observation in Figure~\ref{fig:prompt_cluster}, implying that on the same task, the soft prompts learned by partial PLMs could be potentially transferred to the full-size PLM.

\section{Effect of Partial Model Construction Designs for \ourmodel}
\label{subsec:effect_of_selction}

We construct a partial PLM by dropping a few layers or masking some neurons. As mentioned in \cref{subsec:partial_model_construction}, for layer dropping, we retain the layers uniformly; for FFN reduction, we mask the neurons that are less likely to be activated. How to select the retained parameters is essential to the performance of \ourmodel. To demonstrate this, in Table~\ref{tab:effect_of_selection}, we experiment \ourmodel with another strategy for layer dropping and FFN reduction, respectively.

\begin{table}[!t]
    \centering
    \small
    \begin{tabular}{l@{~~}|c@{~~}|c}
    \toprule[1pt]
        Selection Method & Performance & Relative FLOPs \\
        \midrule[1pt]
        \multicolumn{3}{l}{\textit{Layer Dropping}} \\
        \midrule[1pt]
        Uniform & \textbf{70.92} & \multirow{2}*{72\%}\\
        Last & 69.91 & \\
        \midrule[1pt]
        \multicolumn{3}{l}{\textit{FFN Reduction}} \\
        \midrule[1pt]
        Activation & \textbf{71.50} & \multirow{2}*{84\%} \\
        Random & 66.80 & \\
        \bottomrule
    \end{tabular}
    \caption{Average performance on $5$ investigated tasks using different strategies of layer dropping and FFN reduction on T5$_{\texttt{LARGE}}$.}
    \label{tab:effect_of_selection}
\end{table}

For layer dropping, we compare our strategy of dropping layers uniformly (denoted as \textit{Uniform}) with dropping the last few layers (denoted as \textit{Last}). For both methods, we retain the same number of layers. For example, in order to select $3$ layers from a $24$-layer PLM, the \textit{Uniform} strategy will retain the layer $\{1, 12, 24\}$, and the \textit{Last} strategy will retain the layer $\{1, 2, 3\}$. From Table~\ref{tab:effect_of_selection}, we can derive that the \textit{Uniform} strategy is slightly better than the \textit{Last} strategy. We hypothesize the reason is that the overall functionalities of a PLM 
are uniformly distributed among different layers, and adjacent layers tend to have similar functionalities. Therefore, retaining layers uniformly tends to reserve more functionalities than only retaining the first few layers.

\begin{figure*}[!t]
    \centering
    \includegraphics[width=\textwidth]{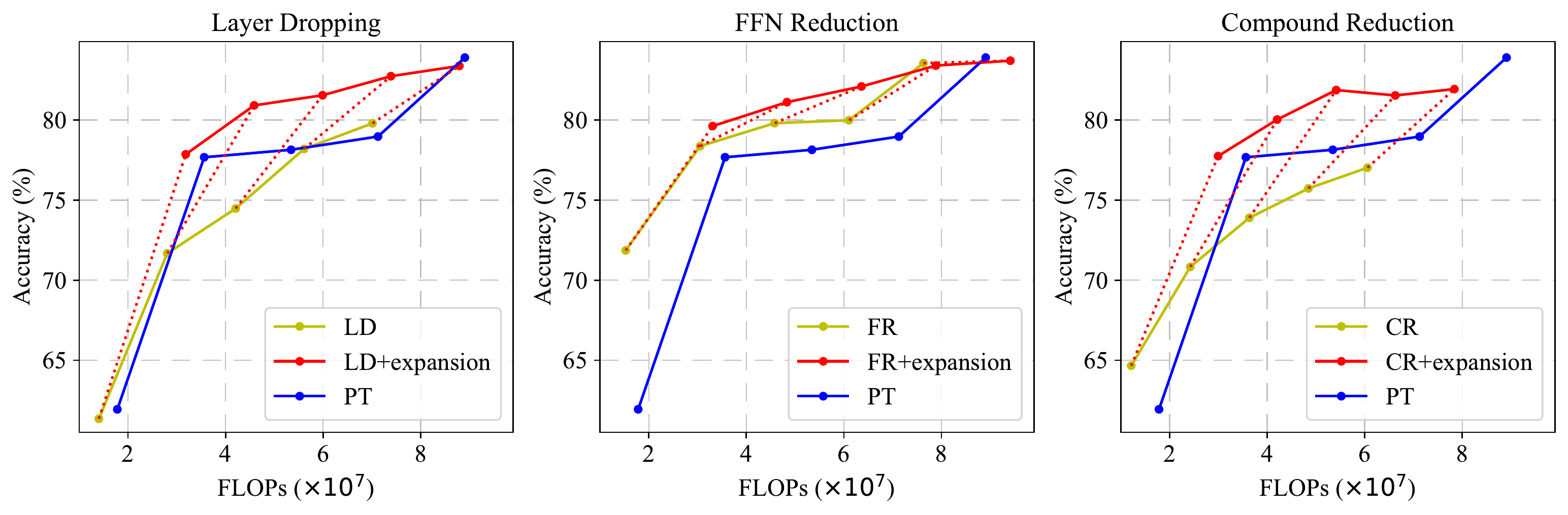}
    \caption{The validation performance on MNLI with different training duration for the last two stages. We conduct this ablation study for each of the three variants of \ourmodel. We compare our \ourmodel with different expanding time (\textcolor[rgb]{1,0,0}{\textbf{red}} line) with vanilla PT (\textcolor[rgb]{0,0,1}{\textbf{blue}} line) and PT without model expansion (\textcolor[rgb]{0.9,0.9,0}{\textbf{yellow}} line).  Each red dot is connected with a yellow dot by a dashed line, indicating it is initialized by the yellow dot and optimized by conducting PT on full-size PLM.}
    \label{fig:effect_of_progressive_steps}
\end{figure*}

For FFN reduction, we compare our strategy of masking neurons based on the activation score (denoted as \textit{Activation}) with randomly masking neurons (denoted as \textit{Random}). For the \textit{Activation} strategy, we feed $1000$ samples prepended by randomly initialized soft prompts into the PLM, and then record the activation score of neurons along each dimension. The results in Table~\ref{tab:effect_of_selection} show that the \textit{Activation} strategy significantly outperforms the \textit{Random} strategy, demonstrating the effectiveness of our method. Randomly masking neurons may abandon those highly activated (most informative) ones, which hinders PT's convergence. We also find empirically the activation score of each neuron in FFN layer may vary a lot across different tasks, which means different neurons may respond differently to the input. This phenomenon also implies that there may exist some ``functional partitions'' in the FFN layers of PLMs.

\section{Effect of Duration for Each Training Stage}
\label{subsec:effect_steps}
To show the effects of the duration of each training stage, following \citet{pmlr-v97-gong19a}, we conduct experiments on MNLI using $\text{T5}_{\texttt{LARGE}}$ with three proposed variants of \ourmodel, and evaluate the effects of training duration for the last two stages.

Specifically, for layer dropping of \ourmodel, we conduct PT on the $18$-layer partial PLM for $15$k steps, and save the learned soft prompts every $3$k steps to get $15 / 3 = 5$ sets of soft prompts. Then using each of these $5$ soft prompts as the initialization, we conduct PT with the full-size PLM for $3$k steps. We report the validation performance and compare \ourmodel with vanilla PT. For FFN reduction and compound reduction of \ourmodel, we conduct similar experiments except that we start from a partial PLM using different construction methods.

The results are shown in Figure~\ref{fig:effect_of_progressive_steps}, from which we can see that expanding the partial PLM's size and then conducting PT (i.e., the red line) performs better than only conducting PT on the partial PLM (i.e., the yellow line). In addition, comparing our \ourmodel (i.e., the red line) with vanilla PT (i.e., the blue line), there is a specific threshold $s'$ of training steps, if we expand the partial PLM before $s'$, the training efficiency can be improved compared with vanilla PT; however, after $s'$, expanding the partial PLM and continuing PT on it does not bring consistent improvement over vanilla PT. In general, expanding the partial PLM between $3$k steps and $12$k steps works well for all three variants of \ourmodel, indicating that within a reasonably broad range, the performance improvement of \ourmodel is not sensitive to the duration of each training stage. We aim to explore how to decide the optimal duration for each training stage in future to make our \ourmodel more practical.

\end{document}